\documentclass{article}

    \usepackage[preprint]{neurips_2024}

\usepackage[utf8]{inputenc} 
\usepackage[T1]{fontenc}    
\usepackage{hyperref}       
\usepackage{url}            
\usepackage{booktabs}       
\usepackage{amsfonts}       
\usepackage{nicefrac}       
\usepackage{microtype}      
\usepackage{xcolor}         

\usepackage{graphicx}
\usepackage{todonotes}
\usepackage{subcaption}
\usepackage{cleveref}
\usepackage{enumitem} 
\usepackage{makecell}
\usepackage{siunitx}

\newcommand{\evtgp}{\textsc{EvtGP}}
\newcommand{\evtphys}{\textsc{EvtPhys}}
\newcommand{\noevtphysclasses}{13}
\newcommand{\noevtgpclasses}{80}
\newcommand{\tilesize}{64x64}

\NewDocumentCommand{\precnum}{mm}{%
  \num[round-mode=places, round-precision=#1]{#2}%
}

\newboolean{show_authors}
\setboolean{show_authors}{true}
\newcommand{\tableprec}{2}

\title{Scalable Geospatial Data Generation Using AlphaEarth Foundations Model}
\author{%
    \ifthenelse{\boolean{show_authors}}
    {
      Luc Houriez\thanks{These authors contributed equally to this work.} \\
      X, the Moonshot Factory, Bellwether\\
      Stanford University \\
      \texttt{houriezl@google.com} \\
      \AND
      Sebastian Pilarski\footnotemark[1] \\
      X, the Moonshot Factory \\
      Bellwether \\
      \texttt{sebpilarski@google.com} \\
      \And
      Behzad Vahedi\footnotemark[1] \\
      X, the Moonshot Factory \\
      Bellwether \\
      \texttt{vahedi@google.com} \\
      \And
      Ali Ahmadalipour \\
      X, the Moonshot Factory \\
      Bellwether \\
      \texttt{aliahma@google.com} \\
      \And
      Teo Honda Scully \\
      X, the Moonshot Factory \\
      Bellwether \\
      \texttt{teonnaise@google.com} \\
      \And
      Nicholas Aflitto \\
      X, the Moonshot Factory \\
      Bellwether \\
      \texttt{aflitto@google.com} \\
      \And
      David Andre\thanks{A complete list of authors and their affiliations is available in the appendix.} \\
      X, the Moonshot Factory \\
      Bellwether \\
      \texttt{davidandre@google.com} \\
    }
    {
      David S.~Hippocampus \\
      Department of Computer Science\\
      Cranberry-Lemon University\\
      Pittsburgh, PA 15213 \\
      \texttt{hippo@cs.cranberry-lemon.edu} \\
      examples of more authors
    }
}

\begin{document}
\maketitle

\begin{abstract}
High-quality labeled geospatial datasets are essential for extracting insights and understanding our planet. Unfortunately, these datasets often do not span the entire globe and are limited to certain geographic regions where data was collected. Google DeepMind’s recently released AlphaEarth Foundations (AEF) provides an information-dense global geospatial representation designed to serve as a useful input across a wide gamut of tasks. In this article we propose and evaluate a methodology which leverages AEF to extend geospatial labeled datasets beyond their initial geographic regions. We show that even basic models like random forests or logistic regression can be used to accomplish this task. We investigate a case study of extending LANDFIRE's Existing Vegetation Type (EVT) dataset beyond the USA into Canada at two levels of granularity: \evtphys\ (\noevtphysclasses\ classes) and \evtgp\ (\noevtgpclasses\ classes). Qualitatively, for \evtphys, model predictions align with ground truth. Trained models achieve 81\% and 73\% classification accuracy on \evtphys\ validation sets in the USA and Canada, despite discussed limitations. 


\end{abstract}
\section{Introduction}
High-quality environmental and climate data is essential for understanding and managing our planet. Nevertheless, comprehensive geospatial datasets that span all regions of the Earth remain rare. This lack of global coverage presents major challenges for applications such as environmental monitoring, disaster prediction, meteorological and climate analysis, and natural resource management.

Traditional geospatial data collection faces significant barriers: acquisition costs are often prohibitive, and integrating multiple datasets requires reconciling vastly different spatial resolutions and temporal sampling rates. While valuable datasets do exist, they are often confined to specific regions or capture only a limited subset of variables needed for comprehensive analysis. Developing reliable methods to extend data coverage into underrepresented areas significantly enhances our ability to monitor earth system changes, design novel applications in otherwise overlooked regions, and make informed predictions at global scales. 

Representation learning models have the potential to fundamentally transform Earth observation capabilities, paralleling the transformative impact of large language models. By training on massive, heterogeneous geospatial datasets, these models learn rich embeddings that encode complex spatiotemporal dynamics and underlying geophysical processes across Earth systems. The key breakthrough lies in their generalization capabilities. They reduce the need for extensive labeled data for specific tasks, as their pre-training allows for more efficient fine-tuning. These learned representations can enable accurate predictions and interpolations of earth system properties in data-sparse regions.

In this paper, we leverage the embeddings generated by Google DeepMind's AlphaEarth Foundations (AEF) model, to address the challenge of data scarcity. AEF generates high-resolution embeddings derived from satellite imagery sources, including Landsat 8 and 9 and Sentinel 1 and 2. A significant advantage of these embeddings is their inherent global coverage and consistent quality. This characteristic makes AEF embeddings particularly well-suited for synthesizing environmental data in otherwise data-deprived areas.

Our core contribution is the development and validation of a pipeline that utilizes the AEF embeddings to generate synthetic environmental datasets. Specifically, we provide a pipeline which trains machine learning models using the globally available AEF embeddings as input features and corresponding ground truth environmental data (e.g., vegetation indices or climatological parameters) as labels. After training, these models can be deployed to infer these same environmental features in areas lacking direct observations by leveraging the AEF embeddings as their input. The global nature of AEF ensures that this methodology is broadly applicable, offering a scalable solution for environmental data augmentation across the planet. 

\subsection{Related Works}
\paragraph{Embedding Models} The utility of learned vector representations or embeddings was first widely adopted in the field of natural language processing with models such as word2vec \cite{mikolov2013}, GloVe \cite{pennington2014}, ELMo \cite{peters2018}, and BERT \cite{devlin2019}. In recent years, there has been a surge of research into the use of embeddings in geospatial applications \cite{mai2023opportunities} with works on models such as place2vec \cite{yan2017}, tile2vec \cite{jean2019}, hex2vec \cite{wozniak2021}, space2vec \cite{mai2020}, GeoVeX \cite{donghi2023}, Terramind \cite{Jakubik2025}, Prithvi \cite{Bodnar2025}, and SatCLIP \cite{klemmer2025satclip}. Most recently, Google DeepMind introduced AlphaEarth Foundations, a multi-modal multi-source geospatial embedding model. This model is exposed to a larger breadth of high quality data sources than previous works and is explicitly trained to be performant across a diverse set of geospatial labeling tasks to make it general purpose \cite{brown2025}. As such, AEF is designed to be effectively used for regression, classification, and segmentation style tasks.

\paragraph{Classification and Segmentation} Researchers have extensively applied machine learning to land cover classification and segmentation, resulting in a rich body of existing literature. Artificial Neural Networks and satellite data were used for land cover classification since the 1990s \cite{hepner1990,civco1993}. Later on, decision-trees \cite{friedl1997,defries1998}, Random Forests \cite{pal2005,gislason2006}, and Support Vector Machines \cite{huang2002,pal2005support} gained traction for land cover classification. Following the advent of deep neural networks in the early 2010s, a variety of such methods along with new sensors and satellite data were implemented for more accurate mapping and classification \cite{scott2017,zhang2018}. In the early 2020s, vision transformers were extensively employed for classification and segmentation of remotely sensed data \cite{scheibenreif2022,yao2023}. A common theme amongst all these works is that they each utilize limited data sources with most limited to one or fewer instruments (e.g. visible, infrared, or SAR data). AEF provides an easy interface into a learned representation spanning many data sources and multiple types of satellite imagery including optical, thermal, and Synthetic Aperture Radar (SAR). As such, it can simplify the process of learning important labels and classifications.

\subsection{AlphaEarth Foundations (AEF)}
Representation learning has proven to be a powerful technique as evidenced in natural language processing and computer vision. The central technique is to utilize encoder models to turn raw data into numerical embeddings. These embeddings are multi-dimensional vectors that capture the semantic essence and spatial relationships within data, thus translating high-dimensional data into lower-dimensional representations to improve data efficiency. Google DeepMind recently released the general-purpose geospatial model AlphaEarth Foundations (AEF) \cite{brown2025}. It learns a dense, global representation of the Earth's surface, capturing intricate patterns and interdependencies. Essentially, the AEF transforms complex Earth observation data into a more structured and dense latent representation, which although less semantically meaningful for humans, provides a structure more amenable for applying ML methods towards downstream tasks.

In order to train the AEF model, the Google DeepMind team leveraged diverse datasets to serve as training inputs and targets. The training inputs consist of Sentinel and Landsat images. The targets consist of various data types including topography (Copernicus DEM), land cover (NLCD), and climate (ERA5-Land). The AEF model uses a self-supervised autoencoder network to learn a representation that enables the reconstruction of individual target datasets from only the input data. It utilizes a novel Space Time Precision (STP) Encoder specifically designed to process long-range relationships across time and space. Learning the final embeddings was accomplished with three neural networks 1) a teacher network that processes complete, unaltered input imagery, 2) a student network that has the same architecture as the teacher network and attempts to produce the same embedding as the teacher network albeit from perturbed or incomplete input data, and 3) a text alignment network that takes text descriptions from wikipedia and produces an embedding. The combination of four loss functions – reconstruction loss, consistency loss, text contrastive loss, and batch uniformity loss across these three networks – produces the final embedding. 

These learned representations are publicly available as a dataset of 64-dimensional vectors for each year, called "Satellite Embedding" dataset \cite{efm_satellite_embedding}. This dataset is a global, annual dataset with a spatial resolution of 10 meters which is currently available from 2017 to 2024.   

\subsection{Problem Statement and Objective}
\paragraph{Problem statement} Raw or information dense data (e.g., satellite or embedding) seldom provides easily extractable insights. Insight extraction usually involves visualizations or transformations to simpler easier-to-understand representations. In geospatial settings, this often takes the form of specified geospatial data labels of important features, e.g., a road or a crop type. Unfortunately, for many applications, labeled data only exists for certain regions. This greatly limits access to model and data interpretability across large swaths of the world.

\paragraph{Objective} The main aim of this article is to investigate whether AEF could be leveraged to generate labels for previously unlabeled regions. Given that the AEF builds on high-quality data sources covering many inputs and modalities, it has the potential to serve as an interpolation base that provides labels with a high-degree of accuracy.

\section{Extending Datasets to New Regions through AEF: A Case Study} 
\label{sec:extending_datasets}

Vegetation type is an important feature humans use when interpreting wildfire risk as it helps understand potential wildfire spread and ignition. Therefore, labeled vegetation information is extremely valuable to explain the outputs of machine-learned wildfire risk models. In this paper, we investigate extending an existing vegetation type dataset from the USA to Canada. This serves as a case study of AEF's potential for dataset extension.

\subsection{Data}
\label{subsec:data}
\paragraph{EVT} LANDFIRE is a multi-agency program run by the US Forest Service, the Department of the Interior and US Geological Survey \cite{rollins2009landfire, la2023landfire}. LANDFIRE provides an ecological dataset called Existing Vegetation Type (EVT) which has historically been used for wildfire management efforts. Developing this dataset required coordinating multiple government agencies to build labeled datasets and training decision tree models to predict existing vegetation type. However, the dataset is limited to the United States (continental, Alaska, Hawaii and insular areas) and serves as a prime example for highly useful data which does not extend globally. In our motivating use case, extending the EVT data to Canada would enable us to make better sense of wildfire risk predictions in Canadian regions.  

\paragraph{EVT granularities} The EVT dataset consists of labels at various levels of classification granularity. LANDIFRE provides label mappings across these different granularity levels, e.g., "Western Hemlock-Yellow-cedar Forest" at medium-level granularity (\evtgp: collapsed vegetation type) maps into "Conifer" at a lower granularity (\evtphys: physiognomy). This paper describes the training results on both \evtphys\ (low) and \evtgp\ (mid) level granularities. 

\begin{figure}
    \centering
    \includegraphics[width=\textwidth]{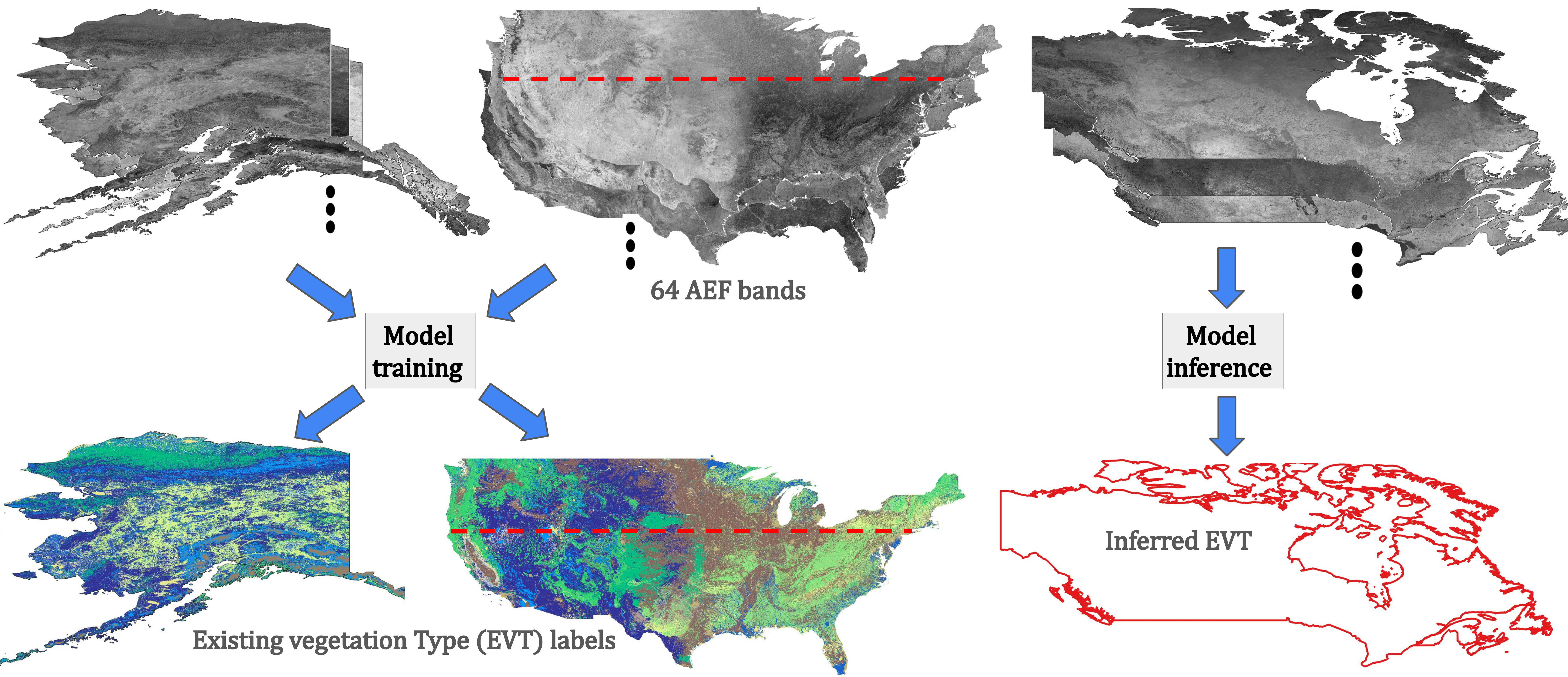}
    
    \caption{Schematic of model training and inference. The 64 bands of AEF data (input) and EVT data (target) from continental USA above the red dotted line and Alaska are used to train the model. Running inference on AEF data in Canada provides expected EVT in the previously unlabeled region.}
    \label{fig:model_training_inference}
\end{figure}

For our model training we set AEF as input and EVT as target output as presented in \Cref{fig:model_training_inference}. Notably, AEF embeddings were evaluated with maximal classification targets of 40 classes. Hence \evtgp\ serves as an investigation into the ability of AEF to generalize to even more difficult classification tasks than originally optimized for. 

\paragraph{Data selection} Deliberate design decisions need to be made when extending a labeled dataset into another region like Canada. In our context, we train on data from Alaska and northern continental US (CONUS) above the 41.6 degrees latitude line as we believe it provides an effective balance between data quantity and regions that exhibit most similar ecological or environmental characteristics to the target Canadian regions. We use LANDFIRE's 2020 release for EVT in our study \cite{la2020landfiredata}.

\paragraph{Data preprocessing} LANDFIRE’s original \evtgp\ classifications consist of 194 unique classes, but we filter out classes comprising less than 0.1\% of the dataset in the training region to address class imbalance. This results in \noevtgpclasses\ classes spanning the selected continental US and Alaska regions. Pixels not belonging to one of these \noevtgpclasses\ are ignored in training via a mask. For \evtphys\ we group all development related classes which yields \noevtphysclasses\ classes from an original 17. We download the AEF embeddings from Google Earth Engine \cite{AEFDataCatalogEE} and train on AEF and EVT with 500m resolutions.

\paragraph{Data splits} We geographically tile our northern CONUS and Alaska data into tiles of size \tilesize. We allocate 90\% of these tiles for training and 10\% for validation. The EVT dataset provides data for a southernmost 90km band of Canada and a western 90km wide band along the Alaska border in recent releases. We reserve this data for our final test set as it directly coincides with our desired target and allows us to evaluate the generalizability of our approach to an unseen region. In all, 4.1 million pixels are used for testing (2.9 million in northern CONUS and 1.2 million in Alaska) and 34.6 million pixels are used for training and validation. 

\subsection{Models}
\label{subsec:models}
Our proposed methodology to extend existing datasets is to train machine learning models to predict labels (e.g., EVT) from AEF inputs. We have developed a flexible and adaptable pipeline wherein model architectures can easily be swapped. This enables rapid experimentation with architectures designed to optimize for certain data characteristics or desired output granularity or resolution. While we apply our methodology and pipeline to extend datasets into Canada from the USA in this article, they are flexible and can be applied to other regions. We evaluate four models for this case study:

\paragraph{Logistic regression}
We train a logistic regression model which for a given AEF pixel determines a linear weighting of the 64 AEF band values to produce an EVT classification \cite{hastie2009elements, scikit-learn}.

\paragraph{Random forest}
We leverage scikit-learn's RandomForestClassifer model to learn the AEF to EVT mapping at the pixel level as above \cite{breiman2001random, scikit-learn}. 

\paragraph{Gradient boosted tree}
We use the LGBMClassifier from the LightGBM library \cite{ke2017lightgbm}. As for the linear and random forest models, individual pixels make up the inputs and outputs. 

\paragraph{Segmentation Model} 
We employ an encoder-decoder semantic segmentation pipeline. It learns by training the encoder and decoder in tandem on an AEF image input and corresponding EVT image target.
The encoder processes the AEF’s representations, extracting and compressing the most relevant signals for the target geospatial feature layer. The decoder reconstructs the label image from this compressed representation. By continuously comparing the model's predicted output against available ground-truth label images during training, the model learns to accurately encode and decode the environmental signals. 

We opted for a fully-convolutional U-Net architecture using EffecientNet-B4 \cite{Tan2019EfficientNetRM} pre-trained on advprop for the encoder (see \Cref{fig:semantic_segmentation} in \Cref{sec:appendix} for architecture details). We train without tile overlap to prevent data leakage to validation sets. During inference to unlabeled regions we set a 50\% overlap on inference tiles to limit border artifacts. Final inference maps are generated by taking output probabilities taken from a final softmax layer and selecting the highest probability class after averaged smoothing across overlaps. We perform training data augmentation consisting of horizontal and vertical flips, random 90 degree rotations, and transposes each with 50\% probability. During training we use Adam as our optimizer, with a learning rate scheduler which reduces on plateau. We train our segmentation model using cross-entropy loss. Training is limited to a maximum of 350 epochs with early stoppage (15 epochs) enabled.

\paragraph{Hardware} The models in this paper are trained on a virtual machine with 160 Intel Broadwell vCPUs and 3844 GB of memory. For the segmentation model, we use a single A100 GPU.

\paragraph{Extending datasets} 
Once training is complete and validated to achieve acceptable metrics, the model is saved. The saved model can then be seamlessly deployed to perform inference in any desired region (including previously unlabeled ones) by running it on AEF inputs of that region. 
Models are appropriately tagged to prevent deployment to regions wherein the target features may fall into completely different distributions as this will likely not yield appropriate labels.

\section{Results}
\label{sec:results}

\begin{figure}
    \centering
    \begin{subfigure}[b]{0.97\linewidth}
        \includegraphics[width=\textwidth]{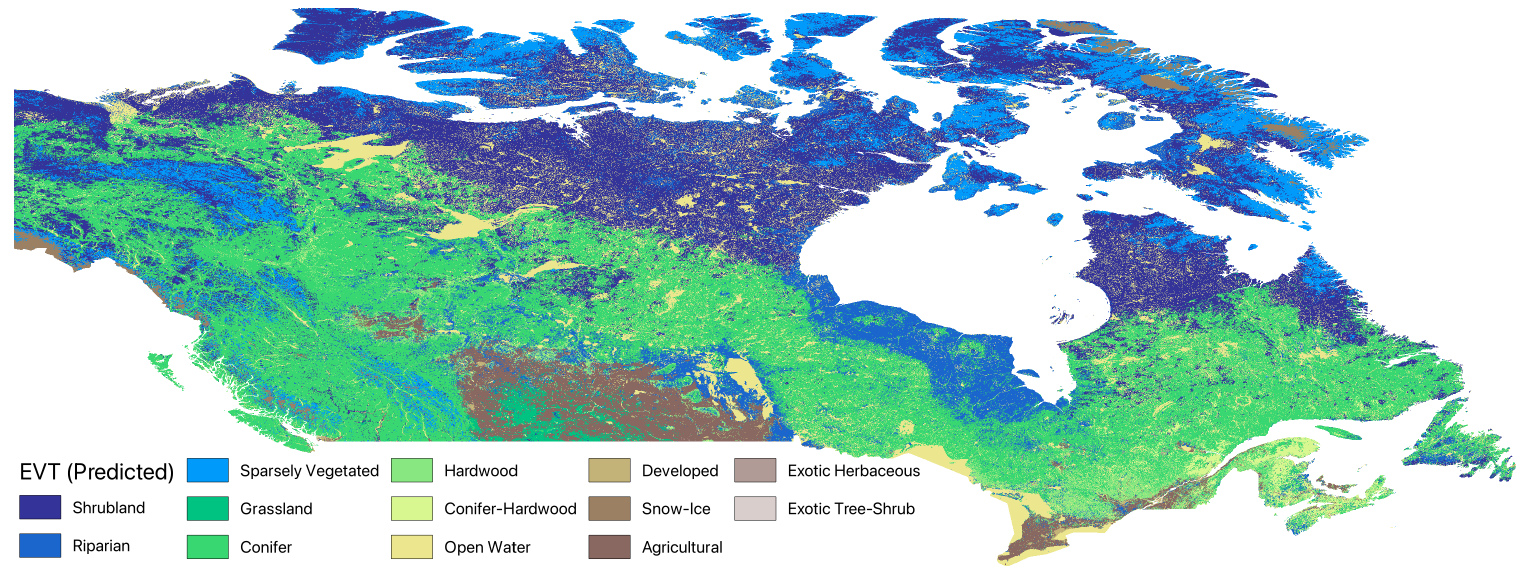}
        \caption{\evtphys\ (\noevtphysclasses\ classes) }
        \label{fig:canada_inference:low_fidelity}
    \end{subfigure}
    \begin{subfigure}[b]{0.97\linewidth}
        \includegraphics[width=\textwidth]{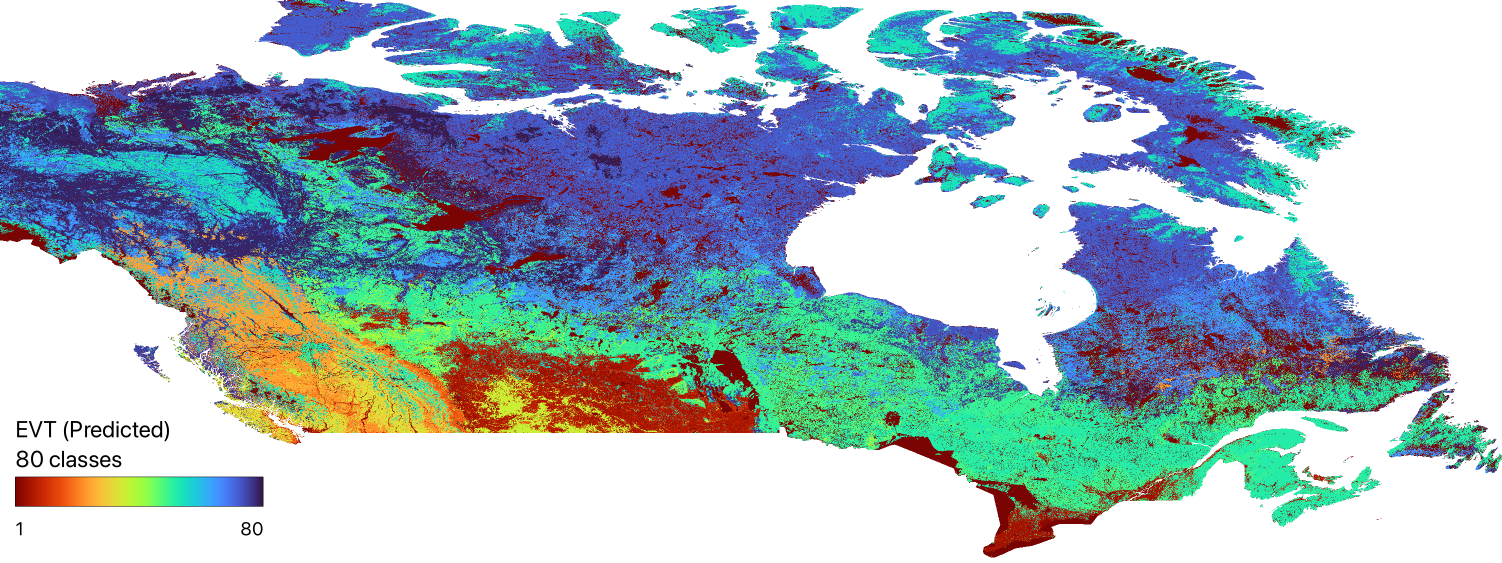}
        \caption{\evtgp\ (\noevtgpclasses\ classes)}
        \label{fig:canada_inference:high_fidelity}
    \end{subfigure}
    \caption{Inference in Canada generated by the segmentation model.}
    \label{fig:canada_inference}
\end{figure}

\paragraph{Qualitative assessment} We present the segmentation model inference maps in Canada for \evtphys\ and \evtgp\ (\Cref{fig:canada_inference}). \evtphys\ inference maps look very similar across all models (\Cref{fig:inference_13}) with differences discussed in \Cref{sec:discussion}. The inference maps seem to exhibit feasible vegetation type continuity. No obvious unexpected vegetation discontinuities or transitions are present. \evtgp\ inference maps generally agree outside northern Canada where significant differences exist. See \Cref{fig:inference_80} for details.

\paragraph{Metrics} 
We quantitatively evaluate the performance of each of our models on both \evtphys\ and \evtgp\ classification granularities using three metrics: Accuracy (ACC), Jaccard Index (J), and F1 score. Both Jaccard Index and F1 are macro-averaged, meaning they are calculated as an unweighted average of the scores for each class. For all three metrics, a higher score indicates better performance. \Cref{tab:results_evtphys} and \Cref{tab:results_evtgp} present results across the training, validation, and test sets defined in \Cref{subsec:data}.

\paragraph{\evtphys} Logistic regression, gradient boosted trees, and the segmentation model achieved similar performance across all metrics for both \evtphys\ training and validation (near 0.79 accuracy, 0.52 Jaccard index, and 0.65 F1 score). The random forest model yielded slightly higher metrics in validation (0.81 accuracy, 0.55 Jaccard, 0.67 F1) and extremely high training values which suggest overfitting (0.97 accuracy, 0.95 Jaccard, 0.97 F1). Significant metric deterioration is observed for all models on the test split. All four models achieved very similar test metrics with random forest besting in all three: 0.73 accuracy, 0.43 Jaccard index, and 0.55 F1 score. Full results are shown in \Cref{tab:results_evtphys}.

\newcommand{\evtphystrnlracc}{0.7713}
\newcommand{\evtphystrnlrj}{0.4788}
\newcommand{\evtphystrnlrf}{0.5959}

\newcommand{\evtphysvallracc}{0.7714}
\newcommand{\evtphysvallrj}{0.4789}
\newcommand{\evtphysvallrf}{0.5949}

\newcommand{\evtphyststlracc}{0.7118}
\newcommand{\evtphyststlrj}{0.3930}
\newcommand{\evtphyststlrf}{0.5050}

\newcommand{\evtphystrnrfacc}{0.9714}
\newcommand{\evtphystrnrfj}{0.9463}
\newcommand{\evtphystrnrff}{0.9717}

\newcommand{\evtphysvalrfacc}{0.8089}
\newcommand{\evtphysvalrfj}{0.5463}
\newcommand{\evtphysvalrff}{0.6711}

\newcommand{\evtphyststrfacc}{0.7327}
\newcommand{\evtphyststrfj}{0.4287}
\newcommand{\evtphyststrff}{0.5462}

\newcommand{\evtphystrngbacc}{0.7934}
\newcommand{\evtphystrngbj}{0.5213}
\newcommand{\evtphystrngbf}{0.6466}

\newcommand{\evtphysvalgbacc}{0.7910}
\newcommand{\evtphysvalgbj}{0.5151}
\newcommand{\evtphysvalgbf}{0.6374}

\newcommand{\evtphyststgbacc}{0.7297}
\newcommand{\evtphyststgbj}{0.4226}
\newcommand{\evtphyststgbf}{0.5400}

\newcommand{\evtphystrnsmacc}{0.7850}
\newcommand{\evtphystrnsmj}{0.5033}
\newcommand{\evtphystrnsmf}{0.6252}

\newcommand{\evtphysvalsmacc}{0.7874}
\newcommand{\evtphysvalsmj}{0.5091}
\newcommand{\evtphysvalsmf}{0.6315}

\newcommand{\evtphyststsmacc}{0.7301}
\newcommand{\evtphyststsmj}{0.4226}
\newcommand{\evtphyststsmf}{0.5418}

\begin{table}[]
\caption{Accuracy (ACC), Jaccard (J), and F1 across data splits for \evtphys\ (\noevtphysclasses\ classes).}
\label{tab:results_evtphys}
\centering
\begin{tabular}{l|lll|lll|lll|}
\cline{2-10}
    & \multicolumn{3}{c|}{\thead{\textbf{Training}}}   & \multicolumn{3}{c|}{\thead{\textbf{Validation}}} & \multicolumn{3}{c|}{\thead{\textbf{Test}}} \\ 
\cline{2-10} 
    & \thead{\textbf{ACC}} & \thead{\textbf{J}} & \thead{\textbf{F1}}    & \thead{\textbf{ACC}} & \thead{\textbf{J}} & \thead{\textbf{F1}}  & \thead{\textbf{ACC}} & \thead{\textbf{J}} & \thead{\textbf{F1}} \\ 
\hline
\multicolumn{1}{|l|}{\textbf{Logistic Regression}}    & \multicolumn{1}{l|}{\precnum{\tableprec}{\evtphystrnlracc}} & \multicolumn{1}{l|}{\precnum{\tableprec}{\evtphystrnlrj}} & \precnum{\tableprec}{\evtphystrnlrf} & \multicolumn{1}{l|}{\precnum{\tableprec}{\evtphysvallracc}} & \multicolumn{1}{l|}{\precnum{\tableprec}{\evtphysvallrj}} & \precnum{\tableprec}{\evtphysvallrf} & \multicolumn{1}{l|}{\precnum{\tableprec}{\evtphyststlracc}} & \multicolumn{1}{l|}{\precnum{\tableprec}{\evtphyststlrj}} & \precnum{\tableprec}{\evtphyststlrf} \\ 
\hline
\multicolumn{1}{|l|}{\textbf{Random Forest}}    & \multicolumn{1}{l|}{\precnum{\tableprec}{\evtphystrnrfacc}} & \multicolumn{1}{l|}{\precnum{\tableprec}{\evtphystrnrfj}} & \precnum{\tableprec}{\evtphystrnrff} & \multicolumn{1}{l|}{\precnum{\tableprec}{\evtphysvalrfacc}} & \multicolumn{1}{l|}{\precnum{\tableprec}{\evtphysvalrfj}} & \precnum{\tableprec}{\evtphysvalrff} & \multicolumn{1}{l|}{\precnum{\tableprec}{\evtphyststrfacc}} & \multicolumn{1}{l|}{\precnum{\tableprec}{\evtphyststrfj}} & \precnum{\tableprec}{\evtphyststrff} \\ 
\hline
\multicolumn{1}{|l|}{\textbf{Gradient Boosted Trees}}    & \multicolumn{1}{l|}{\precnum{\tableprec}{\evtphystrngbacc}} & \multicolumn{1}{l|}{\precnum{\tableprec}{\evtphystrngbj}} & \precnum{\tableprec}{\evtphystrngbf} & \multicolumn{1}{l|}{\precnum{\tableprec}{\evtphysvalgbacc}} & \multicolumn{1}{l|}{\precnum{\tableprec}{\evtphysvalgbj}} & \precnum{\tableprec}{\evtphysvalgbf} & \multicolumn{1}{l|}{\precnum{\tableprec}{\evtphyststgbacc}} & \multicolumn{1}{l|}{\precnum{\tableprec}{\evtphyststgbj}} & \precnum{\tableprec}{\evtphyststgbf} \\ 
\hline
\multicolumn{1}{|l|}{\textbf{Segmentation Model}}    & \multicolumn{1}{l|}{\precnum{\tableprec}{\evtphystrnsmacc}} & \multicolumn{1}{l|}{\precnum{\tableprec}{\evtphystrnsmj}} & \precnum{\tableprec}{\evtphystrnsmf} & \multicolumn{1}{l|}{\precnum{\tableprec}{\evtphysvalsmacc}} & \multicolumn{1}{l|}{\precnum{\tableprec}{\evtphysvalsmj}} & \precnum{\tableprec}{\evtphysvalsmf} & \multicolumn{1}{l|}{\precnum{\tableprec}{\evtphyststsmacc}} & \multicolumn{1}{l|}{\precnum{\tableprec}{\evtphyststsmj}} & \precnum{\tableprec}{\evtphyststsmf} \\ 
\hline
\end{tabular}
\end{table}
\paragraph{\evtgp} In the case of \evtgp, gradient boosted trees and the segmentation model outperformed logistic regression substantially. Gradient boosted trees and segmentation achieve similar metrics for training, but the segmentation model much better generalized to validation and test sets. The segmentation model achieved 0.65 accuracy, 0.28 Jaccard, and 0.40 F1 scores for the training split and similar values for validation. On the test set performance dropped for all models, even more significantly than in \evtphys; the segmentation model had 0.48 accuracy, 0.15 Jaccard, and 0.21 F1. The random forest model again experienced the overfitting phenomenon in training with 0.96 accuracy, 0.94 Jaccard, and 0.96 F1. It outperformed all other models significantly in validation with 0.71 accuracy, 0.39 Jaccard, and 0.53 F1, but performed similarly to segmentation on the test set. Jaccard index and F1 scores are not biased by majority class which, with many classes, likely explains their low values.
Full results are shown in  \Cref{tab:results_evtgp}.

\newcommand{\evtgptrnlracc}{0.5967}
\newcommand{\evtgptrnlrj}{0.2764}
\newcommand{\evtgptrnlrf}{0.4019}

\newcommand{\evtgpvallracc}{0.5964}
\newcommand{\evtgpvallrj}{0.2706}
\newcommand{\evtgpvallrf}{0.3927}

\newcommand{\evtgptstlracc}{0.4199}
\newcommand{\evtgptstlrj}{0.1125}
\newcommand{\evtgptstlrf}{0.1691}

\newcommand{\evtgptrnrfacc}{0.9572}
\newcommand{\evtgptrnrfj}{0.9368}
\newcommand{\evtgptrnrff}{0.9646}

\newcommand{\evtgpvalrfacc}{0.7055}
\newcommand{\evtgpvalrfj}{0.3906}
\newcommand{\evtgpvalrff}{0.5349}

\newcommand{\evtgptstrfacc}{0.4843}
\newcommand{\evtgptstrfj}{0.1637}
\newcommand{\evtgptstrff}{0.2347}

\newcommand{\evtgptrngbacc}{0.6277}
\newcommand{\evtgptrngbj}{0.2948}
\newcommand{\evtgptrngbf}{0.4266}

\newcommand{\evtgpvalgbacc}{0.6195}
\newcommand{\evtgpvalgbj}{0.2799}
\newcommand{\evtgpvalgbf}{0.4058}

\newcommand{\evtgptstgbacc}{0.4420}
\newcommand{\evtgptstgbj}{0.1154}
\newcommand{\evtgptstgbf}{0.1727}

\newcommand{\evtgptrnsmacc}{0.6489}
\newcommand{\evtgptrnsmj}{0.2835}
\newcommand{\evtgptrnsmf}{0.4020}

\newcommand{\evtgpvalsmacc}{0.6581}
\newcommand{\evtgpvalsmj}{0.2938}
\newcommand{\evtgpvalsmf}{0.4113}

\newcommand{\evtgptstsmacc}{0.4813}
\newcommand{\evtgptstsmj}{0.1480}
\newcommand{\evtgptstsmf}{0.2116}

\begin{table}[]
\caption{Accuracy (ACC), Jaccard (J), and F1 across data splits for \evtgp\ (\noevtgpclasses\ classes).}
\label{tab:results_evtgp}
\centering
\begin{tabular}{l|lll|lll|lll|}
\cline{2-10}
    & \multicolumn{3}{c|}{\thead{\textbf{Training}}}   & \multicolumn{3}{c|}{\thead{\textbf{Validation}}} & \multicolumn{3}{c|}{\thead{\textbf{Test}}} \\ 
\cline{2-10} 
    & \thead{\textbf{ACC}} & \thead{\textbf{J}} & \thead{\textbf{F1}}    & \thead{\textbf{ACC}} & \thead{\textbf{J}} & \thead{\textbf{F1}}  & \thead{\textbf{ACC}} & \thead{\textbf{J}} & \thead{\textbf{F1}} \\ 
\hline
\multicolumn{1}{|l|}{\textbf{Logistic Regression}}    & \multicolumn{1}{l|}{\precnum{\tableprec}{\evtgptrnlracc}} & \multicolumn{1}{l|}{\precnum{\tableprec}{\evtgptrnlrj}} & \precnum{\tableprec}{\evtgptrnlrf} & \multicolumn{1}{l|}{\precnum{\tableprec}{\evtgpvallracc}} & \multicolumn{1}{l|}{\precnum{\tableprec}{\evtgpvallrj}} & \precnum{\tableprec}{\evtgpvallrf} & \multicolumn{1}{l|}{\precnum{\tableprec}{\evtgptstlracc}} & \multicolumn{1}{l|}{\precnum{\tableprec}{\evtgptstlrj}} & \precnum{\tableprec}{\evtgptstlrf} \\ 
\hline
\multicolumn{1}{|l|}{\textbf{Random Forest}}    & \multicolumn{1}{l|}{\precnum{\tableprec}{\evtgptrnrfacc}} & \multicolumn{1}{l|}{\precnum{\tableprec}{\evtgptrnrfj}} & \precnum{\tableprec}{\evtgptrnrff} & \multicolumn{1}{l|}{\precnum{\tableprec}{\evtgpvalrfacc}} & \multicolumn{1}{l|}{\precnum{\tableprec}{\evtgpvalrfj}} & \precnum{\tableprec}{\evtgpvalrff} & \multicolumn{1}{l|}{\precnum{\tableprec}{\evtgptstrfacc}} & \multicolumn{1}{l|}{\precnum{\tableprec}{\evtgptstrfj}} & \precnum{\tableprec}{\evtgptstrff} \\ 
\hline
\multicolumn{1}{|l|}{\textbf{Gradient Boosted Trees}}    & \multicolumn{1}{l|}{\precnum{\tableprec}{\evtgptrngbacc}} & \multicolumn{1}{l|}{\precnum{\tableprec}{\evtgptrngbj}} & \precnum{\tableprec}{\evtgptrngbf} & \multicolumn{1}{l|}{\precnum{\tableprec}{\evtgpvalgbacc}} & \multicolumn{1}{l|}{\precnum{\tableprec}{\evtgpvalgbj}} & \precnum{\tableprec}{\evtgpvalgbf} & \multicolumn{1}{l|}{\precnum{\tableprec}{\evtgptstgbacc}} & \multicolumn{1}{l|}{\precnum{\tableprec}{\evtgptstgbj}} & \precnum{\tableprec}{\evtgptstgbf} \\ 
\hline
\multicolumn{1}{|l|}{\textbf{Segmentation Model}}    & \multicolumn{1}{l|}{\precnum{\tableprec}{\evtgptrnsmacc}} & \multicolumn{1}{l|}{\precnum{\tableprec}{\evtgptrnsmj}} & \precnum{\tableprec}{\evtgptrnsmf} & \multicolumn{1}{l|}{\precnum{\tableprec}{\evtgpvalsmacc}} & \multicolumn{1}{l|}{\precnum{\tableprec}{\evtgpvalsmj}} & \precnum{\tableprec}{\evtgpvalsmf} & \multicolumn{1}{l|}{\precnum{\tableprec}{\evtgptstsmacc}} & \multicolumn{1}{l|}{\precnum{\tableprec}{\evtgptstsmj}} & \precnum{\tableprec}{\evtgptstsmf} \\ 
\hline
\end{tabular}
\end{table}
\newcommand{\EvtPhysLrCaAcc}{0.6662} 
\newcommand{\EvtPhysLrCaJ}{0.3074}
\newcommand{\EvtPhysLrCaF}{0.4103}

\newcommand{\EvtPhysLrAkAcc}{0.8198} 
\newcommand{\EvtPhysLrAkJ}{0.3451}
\newcommand{\EvtPhysLrAkF}{0.4185}

\newcommand{\EvtPhysLrUsAcc}{0.5852} 
\newcommand{\EvtPhysLrUsJ}{0.3214}
\newcommand{\EvtPhysLrUsF}{0.4350}

\newcommand{\EvtPhysRfCaAcc}{0.6903}
\newcommand{\EvtPhysRfCaJ}{0.3360}
\newcommand{\EvtPhysRfCaF}{0.4460}

\newcommand{\EvtPhysRfAkAcc}{0.8328}
\newcommand{\EvtPhysRfAkJ}{0.4237}
\newcommand{\EvtPhysRfAkF}{0.5182}

\newcommand{\EvtPhysRfUsAcc}{0.6782}
\newcommand{\EvtPhysRfUsJ}{0.3484}
\newcommand{\EvtPhysRfUsF}{0.4610}

\newcommand{\EvtPhysGbCaAcc}{0.6871}
\newcommand{\EvtPhysGbCaJ}{0.3378}
\newcommand{\EvtPhysGbCaF}{0.4487}

\newcommand{\EvtPhysGbAkAcc}{0.8305}
\newcommand{\EvtPhysGbAkJ}{0.3194}
\newcommand{\EvtPhysGbAkF}{0.3910}

\newcommand{\EvtPhysGbUsAcc}{0.6387}
\newcommand{\EvtPhysGbUsJ}{0.3206}
\newcommand{\EvtPhysGbUsF}{0.4389}

\newcommand{\EvtPhysSmCaAcc}{0.6875}
\newcommand{\EvtPhysSmCaJ}{0.3372}
\newcommand{\EvtPhysSmCaF}{0.4487}

\newcommand{\EvtPhysSmAkAcc}{0.8309}
\newcommand{\EvtPhysSmAkJ}{0.3675}
\newcommand{\EvtPhysSmAkF}{0.4487}

\newcommand{\EvtPhysSmUsAcc}{0.6580}
\newcommand{\EvtPhysSmUsJ}{0.3580}
\newcommand{\EvtPhysSmUsF}{0.4836}

\begin{table}[]
\caption{Model performances for \evtphys\ (\noevtphysclasses\ classes) across 3 distinct test regions. \textit{Canada South} and \textit{Canada West} combined comprise the test set in \Cref{tab:results_evtphys} (see Data Splits in \Cref{subsec:data}).} 
\label{tab:ak_conus_evtphys}
\centering
\begin{tabular}{l|lll|lll|lll|}
\cline{2-10}
    & \multicolumn{3}{c|}{\thead{\textbf{Canada South}}} & \multicolumn{3}{c|}{\thead{\textbf{Canada West}}} & \multicolumn{3}{c|}{\thead{\textbf{Southern CONUS}}} \\ 
\cline{2-10} 
    & \thead{\textbf{ACC}} & \thead{\textbf{J}} & \thead{\textbf{F1}}    & \thead{\textbf{ACC}} & \thead{\textbf{J}} & \thead{\textbf{F1}}  & \thead{\textbf{ACC}} & \thead{\textbf{J}} & \thead{\textbf{F1}} \\ 
\hline
\multicolumn{1}{|l|}{\textbf{Logistic Regression}}  & \multicolumn{1}{l|}{\precnum{\tableprec}{\EvtPhysLrCaAcc}} & \multicolumn{1}{l|}{\precnum{\tableprec}{\EvtPhysLrCaJ}} & \precnum{\tableprec}{\EvtPhysLrCaF}  & \multicolumn{1}{l|}{\precnum{\tableprec}{\EvtPhysLrAkAcc}} & \multicolumn{1}{l|}{\precnum{\tableprec}{\EvtPhysLrAkJ}} & \precnum{\tableprec}{\EvtPhysLrAkF} & \multicolumn{1}{l|}{\precnum{\tableprec}{\EvtPhysLrUsAcc}} & \multicolumn{1}{l|}{\precnum{\tableprec}{\EvtPhysLrUsJ}} & \precnum{\tableprec}{\EvtPhysLrUsF} \\ 
\hline
\multicolumn{1}{|l|}{\textbf{Random Forest}}  & \multicolumn{1}{l|}{\precnum{\tableprec}{\EvtPhysRfCaAcc}} & \multicolumn{1}{l|}{\precnum{\tableprec}{\EvtPhysRfCaJ}} & \precnum{\tableprec}{\EvtPhysRfCaF}  & \multicolumn{1}{l|}{\precnum{\tableprec}{\EvtPhysRfAkAcc}} & \multicolumn{1}{l|}{\precnum{\tableprec}{\EvtPhysRfAkJ}} & \precnum{\tableprec}{\EvtPhysRfAkF} & \multicolumn{1}{l|}{\precnum{\tableprec}{\EvtPhysRfUsAcc}} & \multicolumn{1}{l|}{\precnum{\tableprec}{\EvtPhysRfUsJ}} & \precnum{\tableprec}{\EvtPhysRfUsF} \\ 
\hline
\multicolumn{1}{|l|}{\textbf{Gradient Boosted Trees}}  & \multicolumn{1}{l|}{\precnum{\tableprec}{\EvtPhysGbCaAcc}} & \multicolumn{1}{l|}{\precnum{\tableprec}{\EvtPhysGbCaJ}} & \precnum{\tableprec}{\EvtPhysGbCaF}  & \multicolumn{1}{l|}{\precnum{\tableprec}{\EvtPhysGbAkAcc}} & \multicolumn{1}{l|}{\precnum{\tableprec}{\EvtPhysGbAkJ}} & \precnum{\tableprec}{\EvtPhysGbAkF} & \multicolumn{1}{l|}{\precnum{\tableprec}{\EvtPhysGbUsAcc}} & \multicolumn{1}{l|}{\precnum{\tableprec}{\EvtPhysGbUsJ}} & \precnum{\tableprec}{\EvtPhysGbUsF} \\ 
\hline
\multicolumn{1}{|l|}{\textbf{Segmentation Model}}  & \multicolumn{1}{l|}{\precnum{\tableprec}{\EvtPhysSmCaAcc}} & \multicolumn{1}{l|}{\precnum{\tableprec}{\EvtPhysSmCaJ}} & \precnum{\tableprec}{\EvtPhysSmCaF}  & \multicolumn{1}{l|}{\precnum{\tableprec}{\EvtPhysSmAkAcc}} & \multicolumn{1}{l|}{\precnum{\tableprec}{\EvtPhysSmAkJ}} & \precnum{\tableprec}{\EvtPhysSmAkF} & \multicolumn{1}{l|}{\precnum{\tableprec}{\EvtPhysSmUsAcc}} & \multicolumn{1}{l|}{\precnum{\tableprec}{\EvtPhysSmUsJ}} & \precnum{\tableprec}{\EvtPhysSmUsF} \\ 
\hline
\end{tabular}
\end{table}
\newcommand{\EvtGpLrCaAcc}{0.3262} 
\newcommand{\EvtGpLrCaJ}{0.0706}
\newcommand{\EvtGpLrCaF}{0.1077}

\newcommand{\EvtGpLrAkAcc}{0.6417} 
\newcommand{\EvtGpLrAkJ}{0.2095}
\newcommand{\EvtGpLrAkF}{0.2938}

\newcommand{\EvtGpLrUsAcc}{0.3836} 
\newcommand{\EvtGpLrUsJ}{0.1033}
\newcommand{\EvtGpLrUsF}{0.1563}

\newcommand{\EvtGpRfCaAcc}{0.4028}
\newcommand{\EvtGpRfCaJ}{0.1078}
\newcommand{\EvtGpRfCaF}{0.1554}

\newcommand{\EvtGpRfAkAcc}{0.6770}
\newcommand{\EvtGpRfAkJ}{0.2695}
\newcommand{\EvtGpRfAkF}{0.3698}

\newcommand{\EvtGpRfUsAcc}{0.4798}
\newcommand{\EvtGpRfUsJ}{0.1375}
\newcommand{\EvtGpRfUsF}{0.2059}

\newcommand{\EvtGpGbCaAcc}{0.3629}
\newcommand{\EvtGpGbCaJ}{0.0691}
\newcommand{\EvtGpGbCaF}{0.1032}

\newcommand{\EvtGpGbAkAcc}{0.6289}
\newcommand{\EvtGpGbAkJ}{0.0812}
\newcommand{\EvtGpGbAkF}{0.1138}

\newcommand{\EvtGpGbUsAcc}{0.3023}
\newcommand{\EvtGpGbUsJ}{0.0643}
\newcommand{\EvtGpGbUsF}{0.1050}

\newcommand{\EvtGpSmCaAcc}{0.3985}
\newcommand{\EvtGpSmCaJ}{0.1031}
\newcommand{\EvtGpSmCaF}{0.1479}

\newcommand{\EvtGpSmAkAcc}{0.6771}
\newcommand{\EvtGpSmAkJ}{0.2290}
\newcommand{\EvtGpSmAkF}{0.3105}

\newcommand{\EvtGpSmUsAcc}{0.4556}
\newcommand{\EvtGpSmUsJ}{0.1026}
\newcommand{\EvtGpSmUsF}{0.1577}

\begin{table}[h!]
\caption{Model performances for \evtgp\ (\noevtgpclasses\ classes) across 3 distinct test regions. \textit{Canada South} and \textit{Canada West} combined comprise the test set in \Cref{tab:results_evtgp} (see Data Splits in \Cref{subsec:data}).}
\label{tab:ak_conus_evtgp}
\centering
\begin{tabular}{l|lll|lll|lll|}
\cline{2-10}
    & \multicolumn{3}{c|}{\thead{\textbf{Canada South}}} & \multicolumn{3}{c|}{\thead{\textbf{Canada West}}} & \multicolumn{3}{c|}{\thead{\textbf{Southern CONUS}}} \\ 
\cline{2-10} 
    & \thead{\textbf{ACC}} & \thead{\textbf{J}} & \thead{\textbf{F1}}    & \thead{\textbf{ACC}} & \thead{\textbf{J}} & \thead{\textbf{F1}}  & \thead{\textbf{ACC}} & \thead{\textbf{J}} & \thead{\textbf{F1}} \\ 
\hline
\multicolumn{1}{|l|}{\textbf{Logistic Regression}}  & \multicolumn{1}{l|}{\precnum{\tableprec}{\EvtGpLrCaAcc}} & \multicolumn{1}{l|}{\precnum{\tableprec}{\EvtGpLrCaJ}} & \precnum{\tableprec}{\EvtGpLrCaF}  & \multicolumn{1}{l|}{\precnum{\tableprec}{\EvtGpLrAkAcc}} & \multicolumn{1}{l|}{\precnum{\tableprec}{\EvtGpLrAkJ}} & \precnum{\tableprec}{\EvtGpLrAkF} & \multicolumn{1}{l|}{\precnum{\tableprec}{\EvtGpLrUsAcc}} & \multicolumn{1}{l|}{\precnum{\tableprec}{\EvtGpLrUsJ}} & \precnum{\tableprec}{\EvtGpLrUsF} \\ 
\hline
\multicolumn{1}{|l|}{\textbf{Random Forest}}  & \multicolumn{1}{l|}{\precnum{\tableprec}{\EvtGpRfCaAcc}} & \multicolumn{1}{l|}{\precnum{\tableprec}{\EvtGpRfCaJ}} & \precnum{\tableprec}{\EvtGpRfCaF}  & \multicolumn{1}{l|}{\precnum{\tableprec}{\EvtGpRfAkAcc}} & \multicolumn{1}{l|}{\precnum{\tableprec}{\EvtGpRfAkJ}} & \precnum{\tableprec}{\EvtGpRfAkF} & \multicolumn{1}{l|}{\precnum{\tableprec}{\EvtGpRfUsAcc}} & \multicolumn{1}{l|}{\precnum{\tableprec}{\EvtGpRfUsJ}} & \precnum{\tableprec}{\EvtGpRfUsF} \\ 
\hline
\multicolumn{1}{|l|}{\textbf{Gradient Boosted Trees}}  & \multicolumn{1}{l|}{\precnum{\tableprec}{\EvtGpGbCaAcc}} & \multicolumn{1}{l|}{\precnum{\tableprec}{\EvtGpGbCaJ}} & \precnum{\tableprec}{\EvtGpGbCaF}  & \multicolumn{1}{l|}{\precnum{\tableprec}{\EvtGpGbAkAcc}} & \multicolumn{1}{l|}{\precnum{\tableprec}{\EvtGpGbAkJ}} & \precnum{\tableprec}{\EvtGpGbAkF} & \multicolumn{1}{l|}{\precnum{\tableprec}{\EvtGpGbUsAcc}} & \multicolumn{1}{l|}{\precnum{\tableprec}{\EvtGpGbUsJ}} & \precnum{\tableprec}{\EvtGpGbUsF} \\ 
\hline
\multicolumn{1}{|l|}{\textbf{Segmentation Model}}  & \multicolumn{1}{l|}{\precnum{\tableprec}{\EvtGpSmCaAcc}} & \multicolumn{1}{l|}{\precnum{\tableprec}{\EvtGpSmCaJ}} & \precnum{\tableprec}{\EvtGpSmCaF}  & \multicolumn{1}{l|}{\precnum{\tableprec}{\EvtGpSmAkAcc}} & \multicolumn{1}{l|}{\precnum{\tableprec}{\EvtGpSmAkJ}} & \precnum{\tableprec}{\EvtGpSmAkF} & \multicolumn{1}{l|}{\precnum{\tableprec}{\EvtGpSmUsAcc}} & \multicolumn{1}{l|}{\precnum{\tableprec}{\EvtGpSmUsJ}} & \precnum{\tableprec}{\EvtGpSmUsF} \\ 
\hline
\end{tabular}
\end{table}

\paragraph{Further test set investigation} As metrics for all models are significantly lower on the test set, we investigate it in increased detail and add an additional independent set. First, we separate the original test set (\Cref{subsec:data}) into two independent ones: the southern Canada 90km band and the western Canada 90km band. We also create a test set consisting of CONUS south of the 41.6 degrees latitude line as this was not included in training. Results for \evtphys\ and \evtgp\ are presented in \Cref{tab:ak_conus_evtphys,tab:ak_conus_evtgp}. Model performance varies drastically across these three regions. On \evtphys, all models achieve reasonably similar metric values when measured in Canada South and Canada West. A clear accuracy performance difference exists between Canada South and Canada West, with Canada West exhibiting 0.14 higher accuracy on average across all four models. Generally, Jaccard and F1 scores are also better for Canada West, but exceptions exist such as for gradient boosted trees. Overall, Canada West metrics are much closer to those for the validation set (we discuss possible reasons for this in  \Cref{sec:discussion}: Inference Evaluation and \Cref{fig:groundtruth_vs_pred}). Larger performance differences are observed in Southern CONUS, with random forest and segmentation achieving highest metrics (0.68 accuracy, 0.36 Jaccard, 0.48 F1). This performance difference is amplified for \evtgp, where random forest and segmentation clearly are best across all regions. Random forest achieves highest metric measures across nearly all test regions and metrics for both \evtphys\ and \evtgp.

\section{Discussion}
\label{sec:discussion}

\paragraph{\evtphys\ vs \evtgp\ metrics}
All models trained to predict \evtgp\ (\noevtgpclasses\ classes) achieve significantly lower metric-evaluated performances as compared to models trained for \evtphys\ (\noevtphysclasses\ classes). This is expected as the number of classes and similarity between classes grows. Importantly, these metrics do not consider relative similarity between classes. Notably, even \evtphys\ contains similar classes, e.g., conifer, hardwood, and hardwood-conifer are 3 distinct classes. As can be seen in \Cref{subfig:confusion_matrix}, the conifer-hardwood class is often misclassified as either conifer or hardwood. Such misclassifications are often much more tolerable in practice than the metrics would suggest.

\paragraph{Performance per class}
It is generally expected that model performance will not be the same across all classes. \Cref{fig:class_performance} showcases the segmentation model performance across \evtphys\ classes. Performance does not directly correspond with class quantity in the training set (see \Cref{sec:appendix}). Perhaps unsurprisingly, open water and snow-ice achieve highest precision values. Minority classes with similar more-common classes (e.g., exotic tree shrub, conifer-hardwood, exotic herbaceous) achieve lowest performance scores. Further grouping/clustering would likely significantly improve overall performance and better balance per-class metrics.

\begin{figure}
    \centering
    \includegraphics[width=\linewidth]{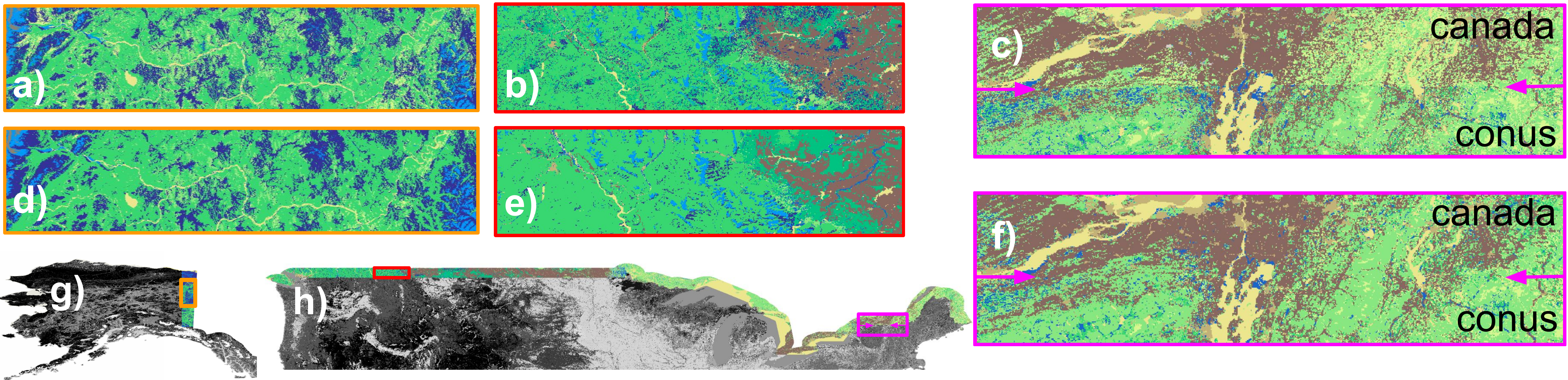}
    \caption{Ground truth \evtphys\ (a–c) compared to gradient boosted trees model inference (d–f) in Canada West (g) and South (h) test regions. Figures (c, f) additionally show land in CONUS across the border which is indicated by the magenta arrows. There, EVT values produced by LANDFIRE seem to exhibit an artificial discontinuity.}
    \label{fig:groundtruth_vs_pred}
\end{figure}

\paragraph{Inference evaluation} 
Most evaluated models experienced similar metric performance across training and validation sets (\Cref{tab:results_evtphys,tab:results_evtgp}). Inference on the test sets, which inherently have different underlying distributions from the training regions, results in significantly lower metrics. Performance across Southern CONUS is likely low due to significant ecological changes in latitudes not present in Alaska and northern CONUS. Notably, we observe that predicted class accuracy drops as a function of distance from labeled training regions (\Cref{tab:latitude_results_distinct}).
As previously noted, models generally achieved higher performance metrics on Canada West than Canada South (\Cref{tab:ak_conus_evtphys,tab:ak_conus_evtgp}). Upon inspection of the ground truth labels, it can be seen in c) of \Cref{fig:groundtruth_vs_pred} that the EVT dataset exhibits an abrupt vegetation change across the CONUS/Canada border which is believed to be artificial, and which isn't seen across the Alaska/Canada border. Across all investigated models, predicted outputs are continuous and do not exhibit such a discontinuity. This suggests that the ground truth label for significant portions of this band include some bias which could lead to lower model measured performance in testing than reality. Nevertheless, all models clearly capture the main \evtphys\ patterns as can be seen on the test tiles of \Cref{fig:groundtruth_vs_pred}. Inference quality drops both qualitatively and quantitatively on the higher granularity \evtgp\ dataset. This is likely the result of increased class quantity and similarity, highlighting the tradeoff between granularity and accuracy. Importantly, we expect that tolerance on accuracy may vary widely depending on the downstream tasks. \evtgp\ also experience performance discrepancies across test regions (\Cref{tab:ak_conus_evtgp}) which can also be partially attributed to the observed EVT discontinuity. 

\paragraph{Random forest performance} Random forest, logistic regression, and gradient boosted trees all managed to achieve very good performance from singular pixel values. This is likely the result of inherent surrounding spatial information encoded into single pixels as a result of AEF being trained on 1.28km neighborhood tiles\footnote{It should be mentioned that this likely results in some minimal leakage of data splits along the tile borders, but should not be significant}. Random forest, surprisingly, outperformed the other models including the segmentation model. This may be due to the fact that EVT labels were generated by decision tree models \cite{la2020landfiredata}. These ground truth labels are not necessarily always correct (output of a predictive model) and they may introduce a exploitable structural bias for a particular model architecture.

\begin{figure}
    \centering
    \includegraphics[width=\linewidth]{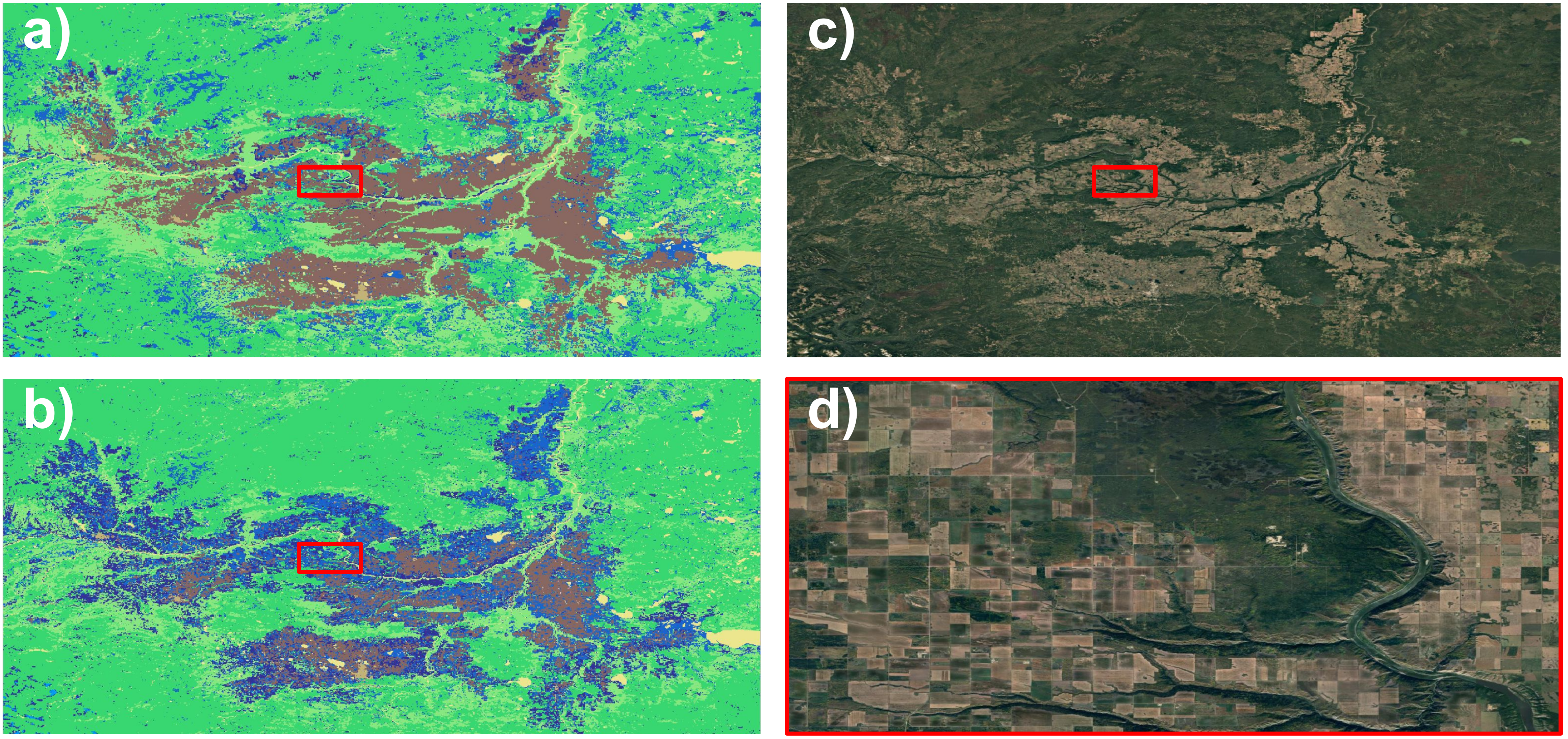}
    \caption{Segmentation model captures farmland (brown \evtphys) better than gradient boosted tree near Peace River in AB, Canada. a) Segmentation model inference, b) Gradient boosted tree model inference, c) Satellite imagery, d) Satellite zoomed-in imagery. }
    \label{fig:grande_prairie}
\end{figure}

\paragraph{Model discrepancies} 
On a zoomed-out prediction map, the majority of predicted pixel classes look consistent between the four evaluated models. One notable observation is that the segmentation model inference results in regions of more consistent vegetation than the other models (thresholding). Other model outputs appear more noisy, with neighboring pixels less likely to belong to the same vegetation class and less clearly defined class boundaries (more akin to LANDFIRE EVT data). This behavior exhibited by the segmentation model doesn't mean it underperforms when comparing its outputs to satellite imagery, quite the contrary. One notable example is near Peace River in AB, Canada (\Cref{fig:grande_prairie}). In this region, the segmentation model correctly identifies (verified by satellite imagery) sections of agricultural land (in brown). Logistic regression, random forests, and gradient boosted trees (pictured) all incorrectly label large sections of the region as shrubland (blue). \Cref{fig:inference_80} in \Cref{sec:appendix} compares all inference maps for \evtgp. Certain discrepancies are visible between the segmentation model and gradient boosted tree, e.g., the darker blue class extends closer to the Canadian border in the gradient boosted tree model. While the examples in this paragraph are handpicked counterexamples and not necessarily representative across the entire datasets, it does point to the possibility that the metrics are not capturing true generalization or performance of respective models.
\paragraph{EVT} Indeed, while we treat EVT as ground-truth in our experimentation, it is inherently noisy as the output of imperfect decision tree models trained on labeled field and satellite data. Misclassifications certainly exist which affect the true metric evaluation of trained models.
AEF opens up new opportunities for potential EVT improvement. A pipeline like the one presented here could be used to train EVT  models using AEF (and possibly other inputs) and LANDFIRE's raw labeled data (target). 
These approaches could possibly lead to better EVT labels across the USA and beyond. A caveat is that AEF is trained on Sentinel data, which limits it to 2017 and onwards.

\section{Conclusion}
This article presents a flexible framework and pipeline that leverages Google DeepMind’s AEF model to extend labeled geospatial data to new regions.  This pipeline trains machine learning model using AEF embeddings as input features and corresponding ground truth environmental data as targets. After training, this model can be deployed to predict labels in previously data-scarce (or missing) regions. We evaluate the approach by using it to extend LANDFIRE’s existing vegetation type (EVT) dataset from the USA to Canada. At a lowest EVT granularity level, \evtphys\ (\noevtphysclasses\ classes), models achieved up to 69\%, 83\%, and 68\% accuracy across Canada South, Canada West, and Southern CONUS test sets. At a higher granularity, \evtgp\ (\noevtgpclasses\ classes), accuracy drops to 40\%, 68\%, and 48\%, respectively. We discuss limitations of measured metrics. Visually and qualitatively, predictions, especially for \evtphys, agree with ground truth classes. 
Given the AEF model is available globally, the work in this paper can be reapplied to other datasets or regions and environmental labels from data-rich to data-scarce regions where similar labels are expected. This opens up tremendous opportunities for improving interpretability of geospatial information across regions.

\bibliographystyle{plain} 
\bibliography{src/bibliography} 
\appendix
\ifthenelse{\boolean{show_authors}}
{
\section{Full Author List and Affiliations}
\label{app:authors} 
\begin{itemize}
    \item Luc Houriez; X, the Moonshot Factory; Bellwether; houriezl@google.com \\Stanford University, Mechanical Engineering Department; houriezl@stanford.com
    \item Sebastian Pilarski; X, the Moonshot Factory; Bellwether; sebpilarski@google.com
    \item Behzad Vahedi; X, the Moonshot Factory; Bellwether; vahedi@google.com
    \item Ali Ahmadalipour; X, the Moonshot Factory; Bellwether; aliahma@google.com
    \item Teo Honda Scully; X, the Moonshot Factory; Bellwether; teonnaise@google.com
    \item Nicholas Aflitto; X, the Moonshot Factory; Bellwether; aflitto@google.com
    \item David Andre; X, the Moonshot Factory; Bellwether; davidandre@google.com
    \item Caroline Jaffe; X, the Moonshot Factory; Bellwether; cjaffe@google.com
    \item Martha Wedner; X, the Moonshot Factory; Bellwether; wedner@google.com
    \item Rich Mazzola; X, the Moonshot Factory; Bellwether; richmazzola@google.com
    \item Josh Jeffery; X, the Moonshot Factory; Bellwether; joshuajeffery@google.com
    \item Ben Messinger; X, the Moonshot Factory; Bellwether; bmessinger@google.com
    \item Sage McGinley-Smith; X, the Moonshot Factory; Bellwether; sagems@google.com
    \item Sarah Russell; X, the Moonshot Factory; Bellwether; sarahrussell@google.com
\end{itemize}
}
{}

\section{Acknowledgments}
The authors wish to acknowledge the valuable input from the Google DeepMind AlphaEarth Foundations team \cite{brown2025} with regards to experiment design and paper review. 

Luc Houriez acknowledges the support of the Stanford Data Science Scholars, advising from Martin Fischer (Stanford University, Civil and Environmental Engineering Department), and Eric Darve (Stanford University, Institute for Computational and Mathematical Engineering). 

\clearpage
\section{Appendix / Supplemental Material}
\label{sec:appendix}

\begin{figure}[h]
    \centering
    \includegraphics[width=0.8\linewidth]{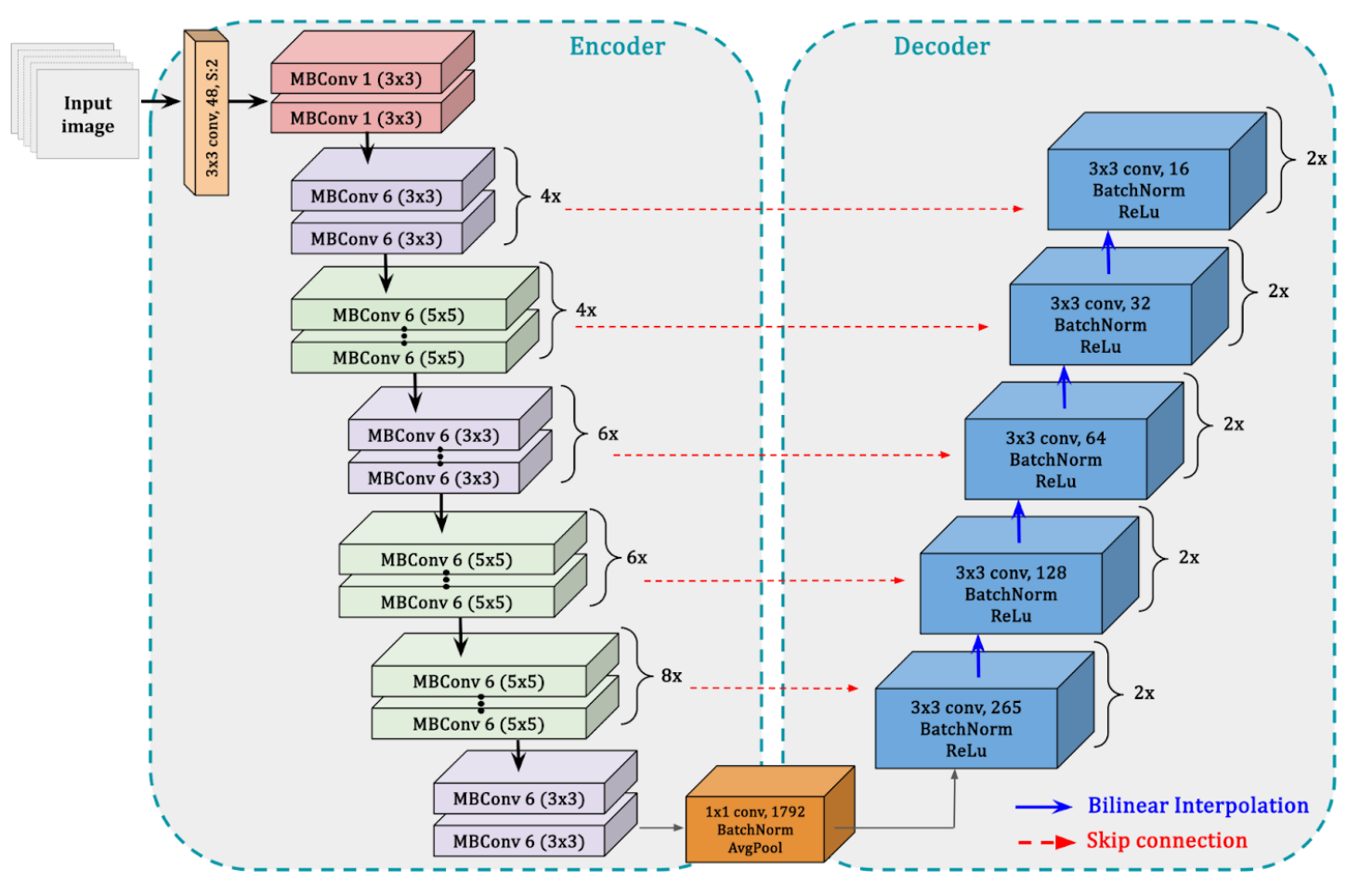}
    \caption{Model architecture. An encoder-decoder semantic segmentation network based on U-Net which uses EfficientNet-B4 as the encoder, and default U-Net decoder.}
    \label{fig:semantic_segmentation}
\end{figure}

\paragraph{Semantic segmentation model}
\Cref{fig:semantic_segmentation} presents the U-Net architecture used in the semantic segmentation model. It consists of EfficientNet-B4 and a default U-Net decoder. EfficientNet-B4 is a convolutional neural network architecture built from MBConv blocks. 

\begin{figure}[h]
    \centering
    \begin{subfigure}[b]{\linewidth}
        \includegraphics[width=\textwidth]{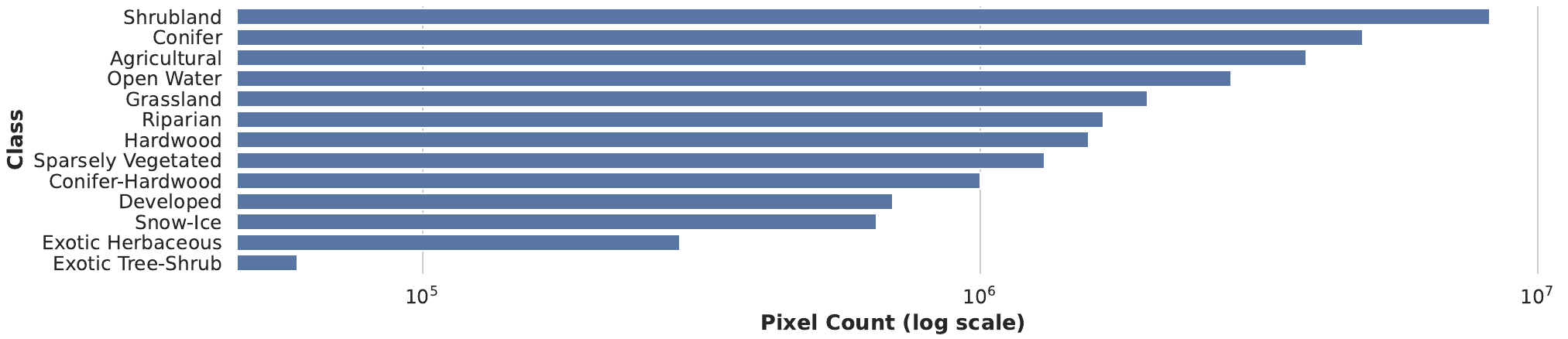}
        \caption{Class distribution for \evtphys\ (13 classes)}
        \label{fig:class_distribution:low_fidelity}
    \end{subfigure}
    \begin{subfigure}[b]{\linewidth}
        \includegraphics[width=\textwidth]{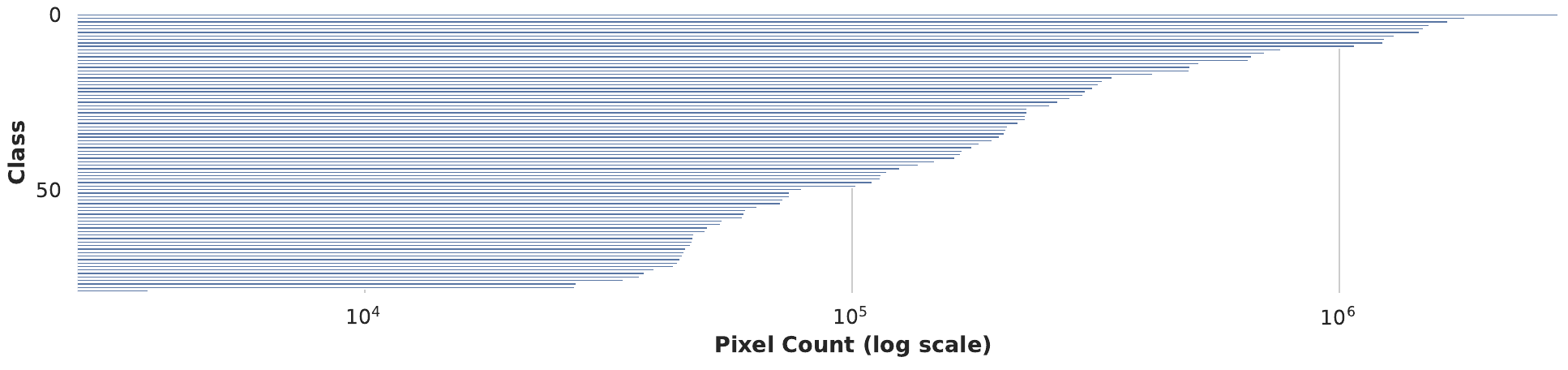}
        \caption{Class distribution for \evtgp\ (80 classes)}
        \label{fig:class_distribution:high_fidelity}
    \end{subfigure}
    \caption{Class distribution in the training data split for both \evtphys\ and \evtgp.}
    \label{fig:class_distribution}
\end{figure}

\paragraph{Class distributions}
We present the class distributions for \evtphys\ (\noevtphysclasses\ classes) and \evtgp\ (\noevtgpclasses\ classes) in log scale in \Cref{fig:class_distribution}. We provide names for the \evtphys\ classes. \evtgp\ class names are omitted due to quantity.

\begin{figure}[h]
    \centering
    \begin{subfigure}[b]{0.49\linewidth}
        \includegraphics[width=\textwidth]{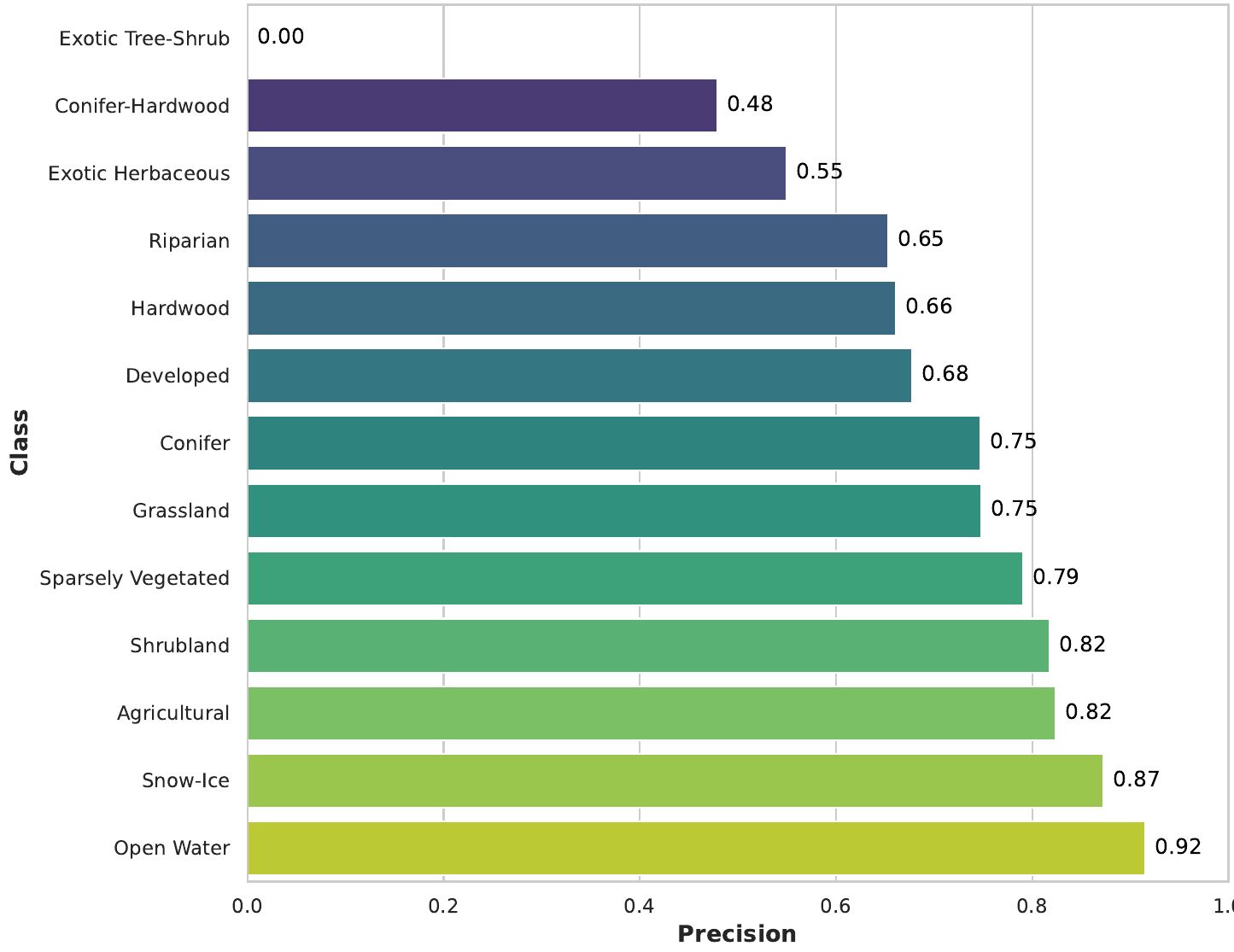}
        \caption{Per-class precision}
        \label{subfig:val_precision}
    \end{subfigure}
    \hfill
    \begin{subfigure}[b]{0.49\linewidth}
        \includegraphics[width=\textwidth]{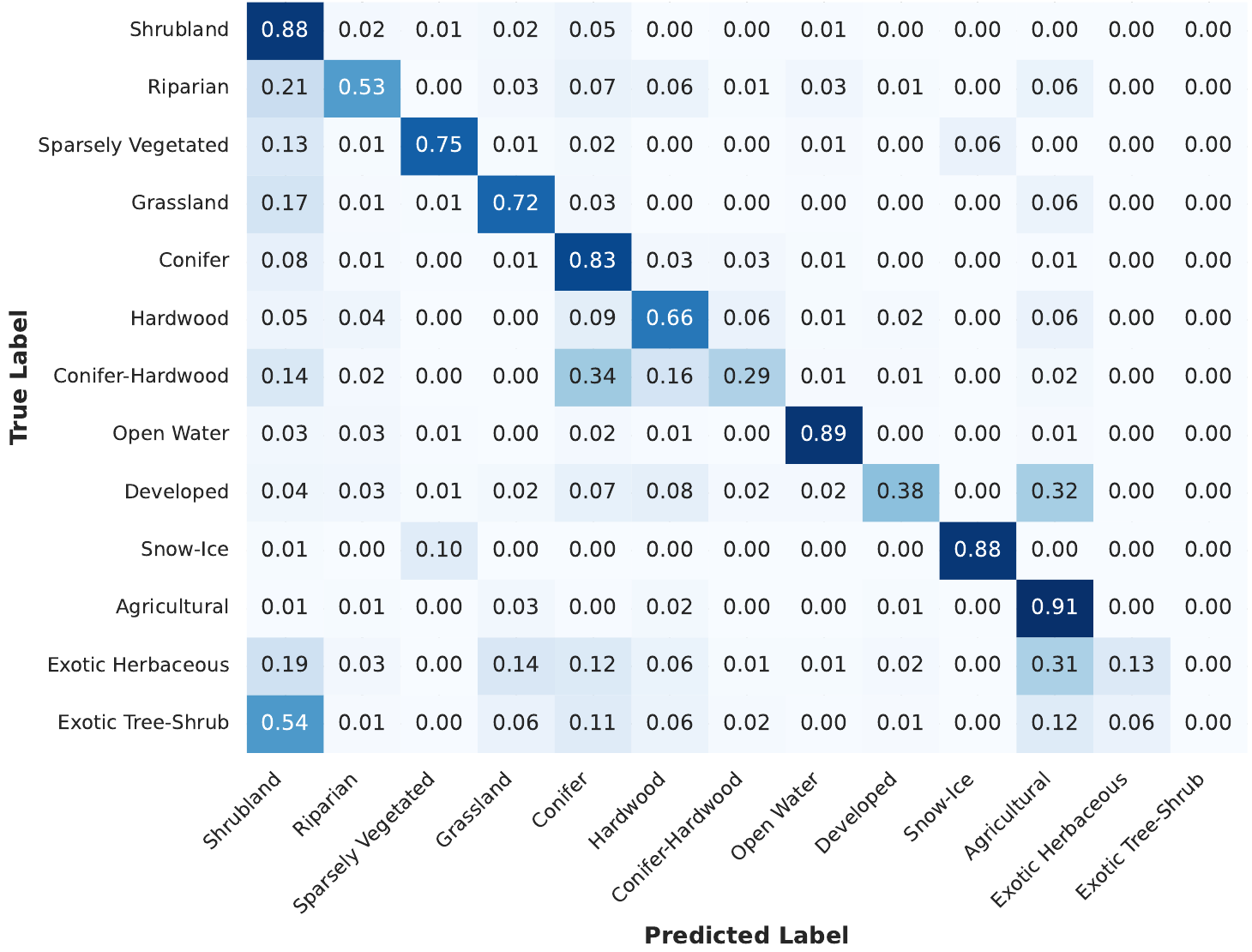}
        \caption{Confusion matrix}
        \label{subfig:confusion_matrix}
    \end{subfigure}
    \caption{Per-class performance by segmentation model for \evtphys}
    \label{fig:class_performance}
\end{figure}

\paragraph{Class performance by segmentation model}
Model performance is not consistent across all classes. In \Cref{fig:class_performance} we present class-specific performance for the segmentation model. Note that precision drops to 0 for the exotic tree shrub as the model misclassifies all instances as just regular shrubland. Random forest achieves 7\% accuracy for this class. Prediction accuracy often drops for less-prevalent classes with similar larger classes.

\paragraph{Performance decreases as a function of distance from the training region} In \Cref{tab:latitude_results_distinct} we presented metric results for gradient boosted trees and segmentation model for \evtphys\ and \evtgp, respectively. All metrics decrease as the geographic distance from the training set (CONUS north of 41.6 degrees of latitude) increases.

\newcommand{\SmEvtgpLatAAcc}{0.5795}
\newcommand{\SmEvtgpLatAJ}{0.1289}
\newcommand{\SmEvtgpLatAF}{0.1909}

\newcommand{\SmEvtgpLatBAcc}{0.4786}
\newcommand{\SmEvtgpLatBJ}{0.0918}
\newcommand{\SmEvtgpLatBF}{0.1384}

\newcommand{\SmEvtgpLatCAcc}{0.3446}
\newcommand{\SmEvtgpLatCJ}{0.0580}
\newcommand{\SmEvtgpLatCF}{0.0888}

\newcommand{\GbEvtphysLatAAcc}{0.7639}
\newcommand{\GbEvtphysLatAJ}{0.4153}
\newcommand{\GbEvtphysLatAF}{0.5275}

\newcommand{\GbEvtphysLatBAcc}{0.6865}
\newcommand{\GbEvtphysLatBJ}{0.3408}
\newcommand{\GbEvtphysLatBF}{0.4523}

\newcommand{\GbEvtphysLatCAcc}{0.5450}
\newcommand{\GbEvtphysLatCJ}{0.2566}
\newcommand{\GbEvtphysLatCF}{0.3698}


\begin{table}[h]
\caption{Test results (Accuracy, Jaccard and F1 scores) for different models across distinct latitude bands within the CONUS region.}
\label{tab:latitude_results_distinct}
\centering
\begin{tabular}{l|lll|lll|lll|}
\cline{2-10}
& \multicolumn{3}{c|}{\thead{\textbf{Lat. 41.6 to 38.6}}} & \multicolumn{3}{c|}{\thead{\textbf{Lat. 38.6 to 35.6}}} & \multicolumn{3}{c|}{\thead{\textbf{Lat. 35.6 to 33.6}}} \\
\cline{2-10}
& \thead{\textbf{ACC}} & \thead{\textbf{J}} & \thead{\textbf{F1}} & \thead{\textbf{ACC}} & \thead{\textbf{J}} & \thead{\textbf{F1}} & \thead{\textbf{ACC}} & \thead{\textbf{J}} & \thead{\textbf{F1}} \\
\hline
\multicolumn{1}{|l|}{\thead[l]{\textbf{Gradient Boosted Trees} \\ \textbf{\evtphys\ (\noevtphysclasses\ classes)}}} & \multicolumn{1}{l|}{\precnum{\tableprec}{\GbEvtphysLatAAcc}} & \multicolumn{1}{l|}{\precnum{\tableprec}{\GbEvtphysLatAJ}} & \precnum{\tableprec}{\GbEvtphysLatAF} & \multicolumn{1}{l|}{\precnum{\tableprec}{\GbEvtphysLatBAcc}} & \multicolumn{1}{l|}{\precnum{\tableprec}{\GbEvtphysLatBJ}} & \precnum{\tableprec}{\GbEvtphysLatBF} & \multicolumn{1}{l|}{\precnum{\tableprec}{\GbEvtphysLatCAcc}} & \multicolumn{1}{l|}{\precnum{\tableprec}{\GbEvtphysLatCJ}} & \precnum{\tableprec}{\GbEvtphysLatCF} \\
\hline
\multicolumn{1}{|l|}{\thead[l]{\textbf{Segmentation Model} \\ \textbf{\evtgp\ (\noevtgpclasses\ classes)}}} & \multicolumn{1}{l|}{\precnum{\tableprec}{\SmEvtgpLatAAcc}} & \multicolumn{1}{l|}{\precnum{\tableprec}{\SmEvtgpLatAJ}} & \precnum{\tableprec}{\SmEvtgpLatAF} & \multicolumn{1}{l|}{\precnum{\tableprec}{\SmEvtgpLatBAcc}} & \multicolumn{1}{l|}{\precnum{\tableprec}{\SmEvtgpLatBJ}} & \precnum{\tableprec}{\SmEvtgpLatBF} & \multicolumn{1}{l|}{\precnum{\tableprec}{\SmEvtgpLatCAcc}} & \multicolumn{1}{l|}{\precnum{\tableprec}{\SmEvtgpLatCJ}} & \precnum{\tableprec}{\SmEvtgpLatCF} \\
\hline
\end{tabular}
\end{table}

\paragraph{Southern CONUS test area} We compare random forest model inference to the ground truth in the southern CONUS test area. Visual agreement seems relatively well achieved, with some notable discrepancies in Texas and New Mexico. 

\begin{figure}[h]
    \centering
        \includegraphics[width=\textwidth]{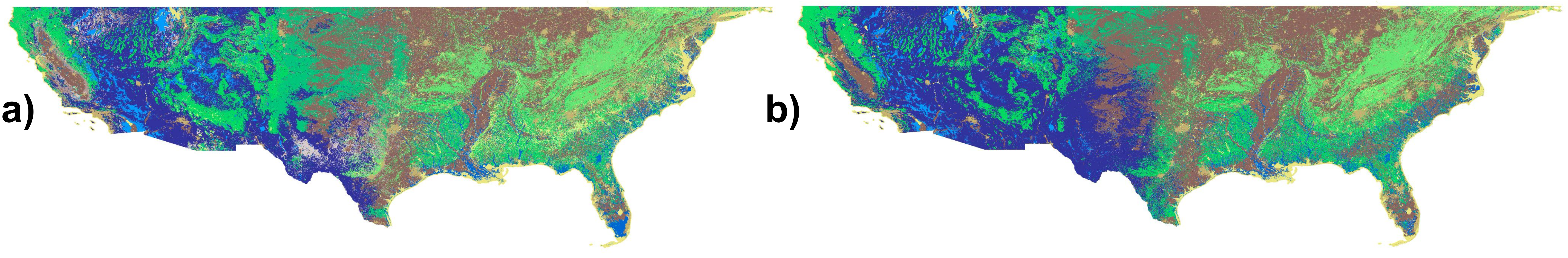}
    \caption{Southern CONUS test area (below latitude 41.6). \evtgp\ ground truth (a) versus inference using random forest (b)}
    \label{fig:southconustest}
\end{figure}

\paragraph{Comparing model inferences} We compare model inferences for all models for \evtgp\ and \evtphys. Overall, good consistency is observed. Inference from logistic regression looks quite different for \evtgp.

\begin{figure}[h]
    \centering
        \includegraphics[width=\textwidth]{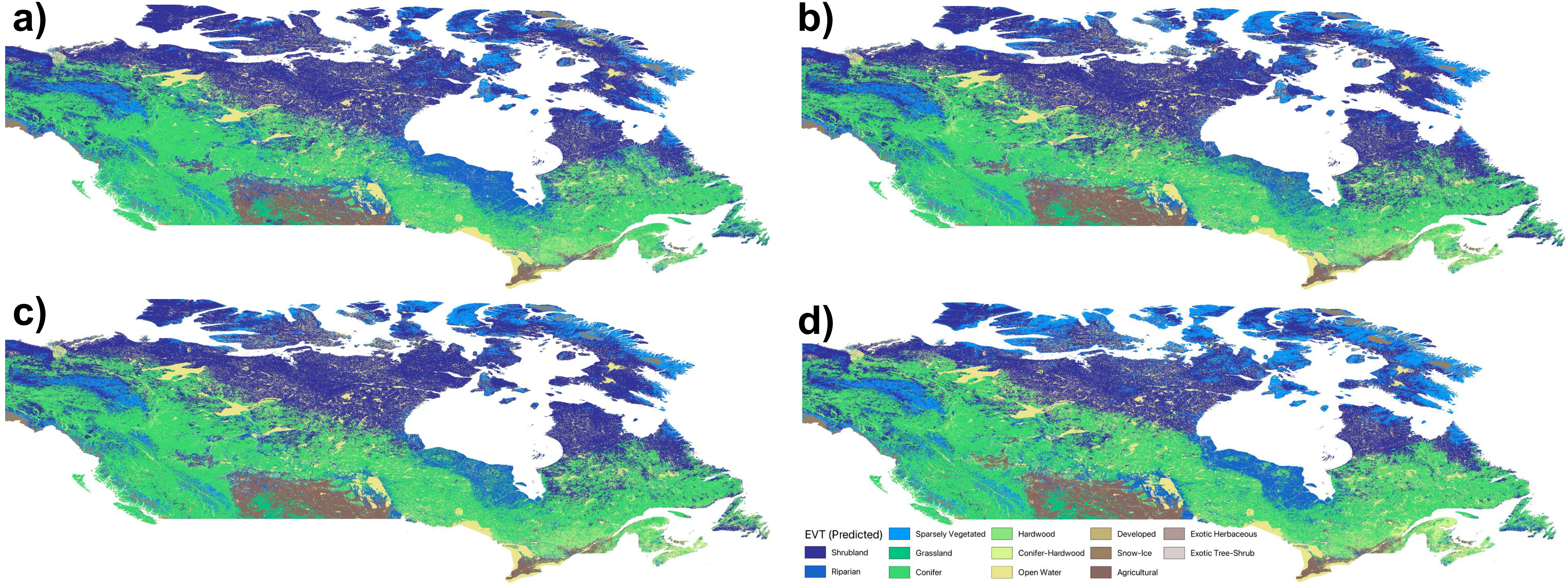}
    \caption{\evtphys\ (13 classes) inference in Canada  using: (a) logistic regression, (b) random forest, (c) gradient boosted trees and (d) segmentation model. }
    \label{fig:inference_13}
\end{figure}

\begin{figure}[h]
    \centering
        \includegraphics[width=\textwidth]{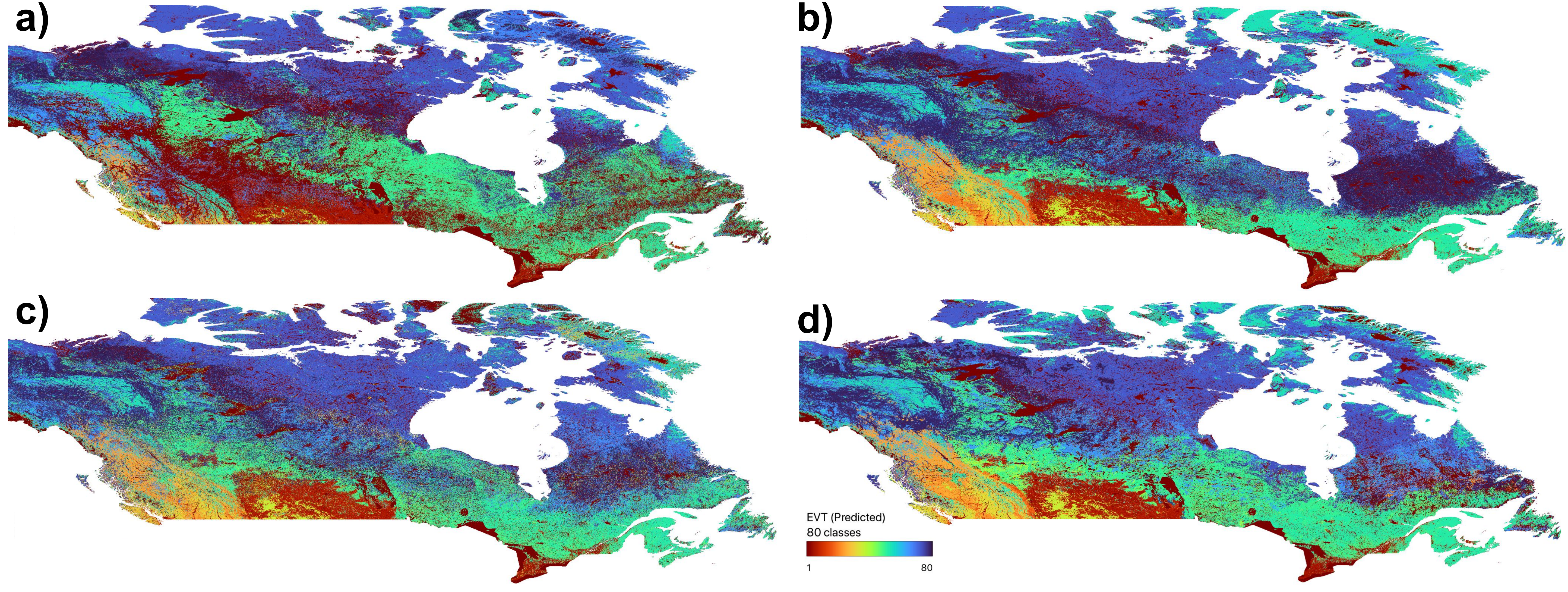}
    \caption{\evtgp\ (80 classes) inference in Canada using: (a) logistic regression, (b) random forest, (c) gradient boosted trees and (d) segmentation model. }
    \label{fig:inference_80}
\end{figure}


\end{document}